# A Diverse Clustering Particle Swarm Optimizer for Dynamic Environment: To Locate and Track Multiple Optima

Zahid Iqbal, Waseem Shahzad *Member, IEEE*

*Abstract*— In real life, mostly problems are dynamic. Many algorithms have been proposed to handle the static problems, but these algorithms do not handle or poorly handle the dynamic environment problems. Although, many algorithms have been proposed to handle dynamic problems but still, there are some limitations or drawbacks in every algorithm regarding diversity of particles and tracking of already found optima. To overcome these limitations/drawbacks, we have proposed a new efficient algorithm to handle the dynamic environment effectively by tracking and locating multiple optima and by improving the diversity and convergence speed of algorithm. In this algorithm, a new method has been proposed which explore the undiscovered areas of search space to increase the diversity of algorithm. This algorithm also uses a method to effectively handle the overlapped and overcrowded particles. Branke has proposed moving peak benchmark which is commonly used MBP in literature. We also have performed different experiments on Moving Peak Benchmark. After comparing the experimental results with different state of art algorithms, it was seen that our algorithm performed more efficiently.

*Index Terms*—Classical Optimization Techniques, Clustering, Particle Swarm Optimizer, Dynamic Environment, Diversity, tracking optima

## I. Introduction

A lot of work has been done and is in progress in EAs for static environment [1]. But, recently, research in EAs is taking turn towards dynamic environment because most of the real world problems which we face in daily life are dynamic.in static environment, no record is maintained for the changing optima instead we just locate the optima but in dynamic environment, we have to not only locate the multiple optima but also have to track the trajectory of changing optima in search space.

Several techniques in evolutionary schemes have been developed to address dynamic optimization problems. These schemes include diversity schemes [2] [3] [4], memory schemes [5] [6] [7] multi-population schemes [8] [9] adaptive schemes [10] [11] [12] multi-objective optimization methods [13] [14] [15] and problem change detecting approaches [16] and prediction schemes [17].

PSO has been very commonly used due to its attractive features like ease of implementation and no gradient information needed. It is being used to solve a large number of optimization problems although some of these problems can also be solved by genetic algorithms, neural networks etc. So, it can be used where we do not have gradient or where usage of gradient can be costly.

Basic PSO is a stochastic optimization technique which has been performed well for static optimization problems [18]. But diversity loss is the main limitation of basic PSO to handle dynamic environment. The gbest particle plays a critical role to reduce the diversity of the basic PSO. It strongly attracts the other particles, so they prematurely converge on the local optima or global optima towards the gbest particle. Whereas for DOPs, these particles should locate more and more local optima so that they may help in next environment. Diversity can be increases by using multi-population method so in this method multiple clusters cover different peaks. But, then, we have to face the problems like how to guide the particles to move towards a sub region, how to determine number of particles in a cluster, how to determine the number of clusters and so on.

Recently, a clustering PSO has been proposed by Yang and Li to handle these problems in [19]. They have used the nearest neighbor learning strategy for training the particles and hierarchal clustering method to locate and track multiple optima in dynamic environment. They employ the clustering in two phases

1) rough clustering and
2) refining clustering.

Although in [20], they have simplified their approach by removing the training method and employing the clustering method in one step. Because instead using refining clustering, same objective can be achieved by using a threshold *max_subsize* which controls the number of particles in a cluster. Second, training process was removed because there was no further need of it in [20] so in this way, computational resources can be used for local search.

At last, in experimental study section, different comparisons have been performed between our proposed method and recently published algorithm CPSO.

Zahid Iqbal was with Department of IT in university of Punjab. Now he is with Faculty of Computing and IT in University of Gujrat, Punjab, Pakistan; (e-mail: zahid.iqbal@uog.edu.pk).
Waseem Shahzad is with National University of Computer and Emerging Sciences. (waseem.shahzad@nu.edu.pk)

## II. Related Work

Basic algorithm can be describes in the following way, a swarm of particles fly to find the peaks in search space. Each particle keeps record of its own best position in the search space, which is called pbest position of a particle. Each particle also keeps record of best value, called lbest value, which is the best value among all pbest values of all particles. In each iteration, each particle updates its velocity and position and move towards the lbest. The velocity equation has social component as well as cognitive component. Setting a small social component and large cognitive component may help the particles to exploit the search space effectively and to avoid to stuck in local minima. Whereas setting a large social component and small cognitive component may help the particles to quickly converge on global optima. Different versions of PSO have been proposed so different variations of velocity equation have been proposed. In [21] velocity and position of particles is updated in the following way.

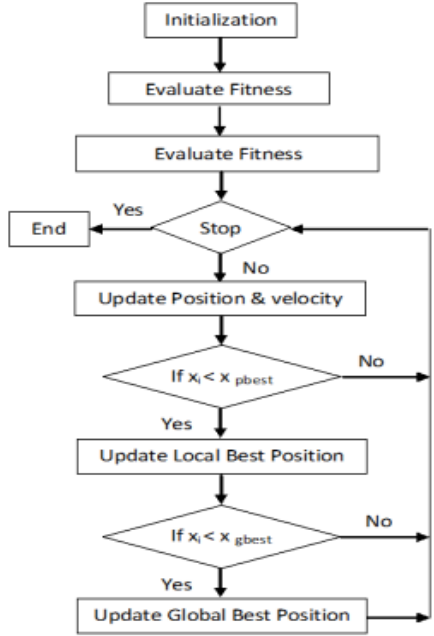

Fig 1. Basic PSO

$$v_{id}(t+1) = \gamma v_{id}(t) + \dot{\eta}_1 r_1(pbest_d - x_{id}(t)) + \dot{\eta}_2 r_2(gbest_d - x_{id}(t)) \quad (1)$$

$$x_{id}(t+1) = x_{id}(t) + v_{id}(t+1) \quad (2)$$

Here

$$\dot{\eta}_1 r_1(pbest_d - x_{id}(t))$$

is called cognitive component and

$\dot{\eta}_2 r_2(gbest_d - x_{id}(t))$ is called social component. Where

w : inertial weight .

$\eta_1, \eta_2$: acceleration constants.
$r_1, r_2$: random numbers.
$x^d_i(t)$ : position of particle i in dimension d at time t.
$x^d_i(t+1)$: position of particle in dimension d at time t+1.
$v^t_i$: velocity of particle i in dimension d at time t.
$v^d_i(t+1)$ : velocity of particle i in dimension d at time t+1.
$pbest^d_i$: pbest of particle i in dimension d.
$gbest^d$: gbest of the group.

$r_1 \in U(0,1)$ and $r_2 \in U(0,1)$ in combination with

$$0 < c_1, c_2 \geq 2$$

are used to determine the maximum step that a particle can take in each iteration. $w \in U(0,1)$ determines the effect of the previous velocity on new calculated velocity.

The framework of the original PSO is shown in Basic PSO algorithm. According to Kennedy, if we remove the social component from the equation (2) i.e. use of the cognition only model degrades the performance of the swarm [22]. This may be due to fact that there is no interaction between different particles. But when we use the social only model then it gives the performance better than original PSO on specific problems.

Basically, there are two main versions of the PSO: *lbest model* and *gbest model*. They are categorized on their neighborhood. In gbest model, neighborhood consists of whole swarm so each particle can share information with any other particle whereas in lbest model, neighborhood consists of only some fixed particles. Both models give different performance on different problems. According to Kennedy and Eberhart [22] and Poli et al. [23], gbest model results in faster convergence but have a high probability of getting stuck in local optima. Whereas lbest model have less chances of getting stuck in local optima but also have slow convergence speed.

## III. Proposed Technique

Our algorithm works in the following way: Algorithm starts by generating a cradle swarm. After generating clusters using single linkage hierarchal clustering method, these clusters exploit those sub-regions of search space which are covered by these clusters. Then overlapping, overcrowding and convergence check are performed on these clusters to control the redundancy. At the end of each iteration, if an environmental change is detected then a new cradle swarm is generated with reservation of positions of converged clusters of previous environments.

In cluster based PSO algorithms, number of clusters and size of clusters play an important role. If we distribute too many clusters in search space, then it may be wastage of computational resources. On the other hand, if there are too small clusters then algorithm may not locate the peaks efficiently. To overcome this problem, different methods have been used to generate multiple clusters e.g. single linkage hierarchal clustering : Shengxiang Yang and Changhe Li '10 [20], k-means clustering algorithm: Kennedy'00 [9], shifting balance GA (SBGA): Oppacher and Wineberg '99, Self-organizing scouts (SOS) GA: Branke et al '00, nbest PSO and niching PSO (NichePSO): Brits '02, Speciation based PSO (SPSO): Parrott and Li '04 [23], Charged PSO (mCPSO) and quantum swarm optimization (mQSO): Blackwell and Branke '06 [25]. Each clustering method has different pros and cons. For example, number of clusters must be predefined in k-means, mCPSO and mQSO. But it is very difficult to know the optimum value of k for a specific population and it is usually problem dependent. Setting non-optimum value of k can result in improper number of clusters. For SPSO, mCPSO, and mQSO, the search radius must be given by experimental results. Similarly, NichePSO and SPSO, create the clusters without analyzing the distribution of population. We are following the Single Linkage Hierarchal Clustering method. We have compared this with other methods of creating multiple clusters and found that it is more suitable to automatically create proper number of clusters in different areas of search space. This method works for any dynamic environment especially for undetectable dynamic environments.

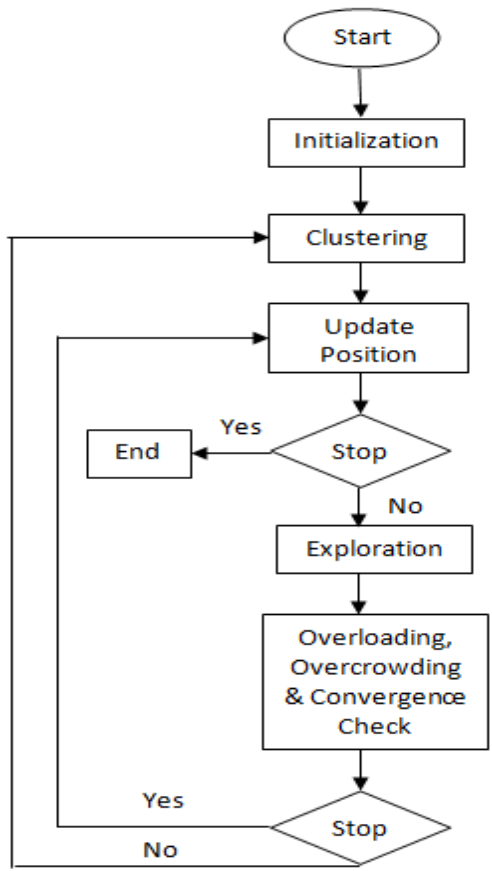

Fig 2. Flow Chart of DCPSO

The clustering method works in following way. First a list G is created of size cradle swarm in such a way that each cluster in G contains just one particle of cradle swarm. Then *FindNearestPair* algorithm is called iteratively to find a such pair of clusters which have shortest distance between them and total number of particles in both clusters does not exceed by a specific threshold (max subsize) .Once such pair has been found then these two clusters are merged together. This process continues until all clusters in G have particles more than 1. By using the max subsize threshold, number of clusters and size of each cluster is automatically determined. As described in algorithm PSO.

A linear decreasing scheme has been also used for inertia *w* to fast the convergence process as in [21].

$$w = w_{max} - \{(w_{max} - w_{min}) * c\_itr\} / r_{itr} \qquad (3)$$

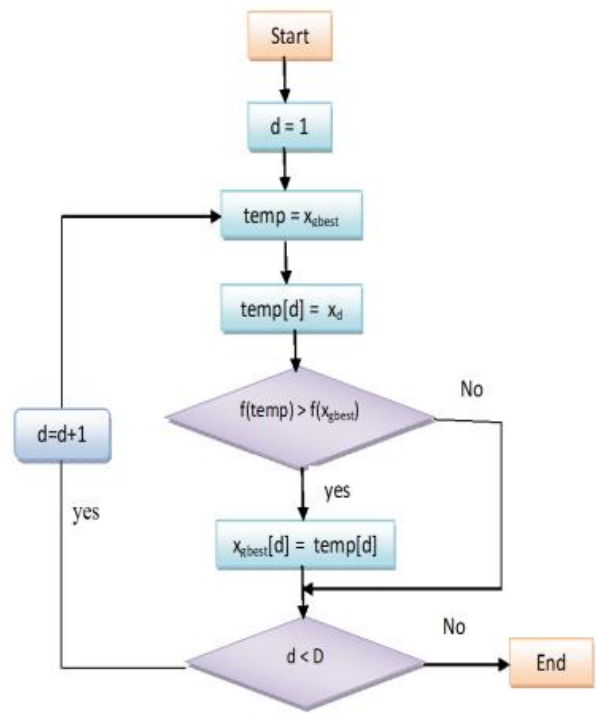

Fig 3. Updation of gbest of a cluster

Here,
 $w_{max}$ = upper bound of inertia
 $w_{min}$ = lower bound of inertia
 $c\_itr$ = count cluster iteration,
 $r_{itr}$ = number of iterations remaining before environment change

To calculate the remaining iterations
$$r_{itr} = (U_{cf} - evals) / p\_size$$
where
 $U_{cf}$ = change frequency i.e. environment will change after these fitess evaluations
 $p\_size$ = total particles in all clusters including cradle swarm

We have used a different method to update local best of each cluster during exploration [20]. In traditional way, when a lbest position of a cluster is updated then all dimensions of this lbest are replaced with new position. In this way, some dimensions may have promising information which will be lost due to updating of all dimensions of lbest. To overcome this problem, useful information is extracted from improved dimensions of improved particles rather than all dimensions of a lbest are updated. When a new position is found, each dimension of lbest is check iteratively. We replace the dimension of lbest with corresponding dimensional value of particle if a better fitness value is found. After updating the pbest and lbest of particles of all clusters, a new technique is used to update the lbest of worst cluster among other clusters as shown in Algorithm 2. A worst cluster is identified based on the fitness of its lbest position. First a worst cluster is identified. We extract useful information from best dimensions of lbest of other clusters. We iteratively update the dimensions of lbest of worst cluster, if we found better fitness value against that position. Once lbest of worst cluster has been updated then all particles of worst clusters are moved to that updated lbest.

We can, also, use the concept of confidence value i.e. if we found a new position; we set its confidence value to 1 and if, again, this position comes then its confidence value will be incremented and so on. If we get a position with confidence value higher than or equal to 2 then we move the particles of worst cluster to that new position.

Normally overlapping check is performed based on the radius of the cluster. Distance between lbest of clusters is calculated and if this distance is less than the radius of the clusters then they are combined or any one is removed. In this method, it is assumed that each cluster just covers one peak. But, in real scenario, a cluster can cover more than one peak so the merging or removing the cluster is not an efficient method.

To check overlapping of different clusters, we have used a method defined in [20]. In this method, first we find two overlapping clusters and calculate the overlapping ration between these two clusters. To calculate ratio of cluster *a*, we find number of particles of cluster a within the radius of cluster *b* and vice versa. After calculating the percentage, smaller percentage is taken and if this percentage is less than a specific threshold *R* then both clusters are merged together. We do this process for all clusters.

After handling the overlapped clusters, overcrowding check is performed otherwise too many particles may search for a single peak so this can be wastage of resources. In this method, we remove a specific numbers of particles from a cluster if number of particles it contained bypass a specific threshold. Here *max_subsize* is a threshold value.

A convergence check is performed to identify whether a cluster has converged or not. For this purpose, we set a small threshold value. If the radius of a cluster is less than this small threshold value, then we consider that cluster as a converged cluster. We save its lbest to track the movements in next environment.

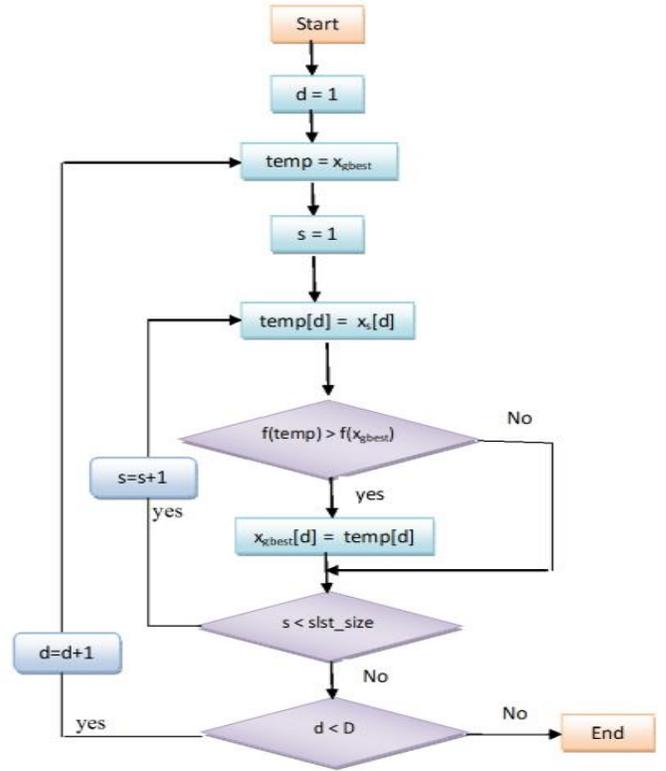

Fig 4. Finding a new best position

When we perform convergence and overlapping check then there can be a scenario that we are with no clusters. So, to handle this specific case, we repopulate the main cluster with some random particles.

Every dynamic PSO must be able to detect the environmental changes [26]. There are different approaches to detect environmental change. For example, we can use the deterioration in population performance or time averaged best performance [27]. We may use some monitoring particles in fitness space. We can iteratively check the fitness of these particles. If fitness of these particles changes then we can say that an environmental change has occurred. We can also reevaluate the pbest of each particle before updating [23]. Similarly, there are other approaches too. In this paper, we are using lbest positions to detect environmental change. We simply reevaluate the lbest particles over all clusters. If an environmental change is detected, we save the lbest of each remaining clusters. Then add these saved lbest into a new generated main cluster. Then again clustering is performed on the new generated cluster.

## IV. EXPERIMENTAL STUDY

The default configurations which we have used in our proposed algorithm are given in Table 1. We have performed the experiments on MPB problem proposed by Branke because it is the widely used dynamic benchmark. In this moving peak benchmark problem, a peak can change its height, width, and location. This MPB problem can be defined as:

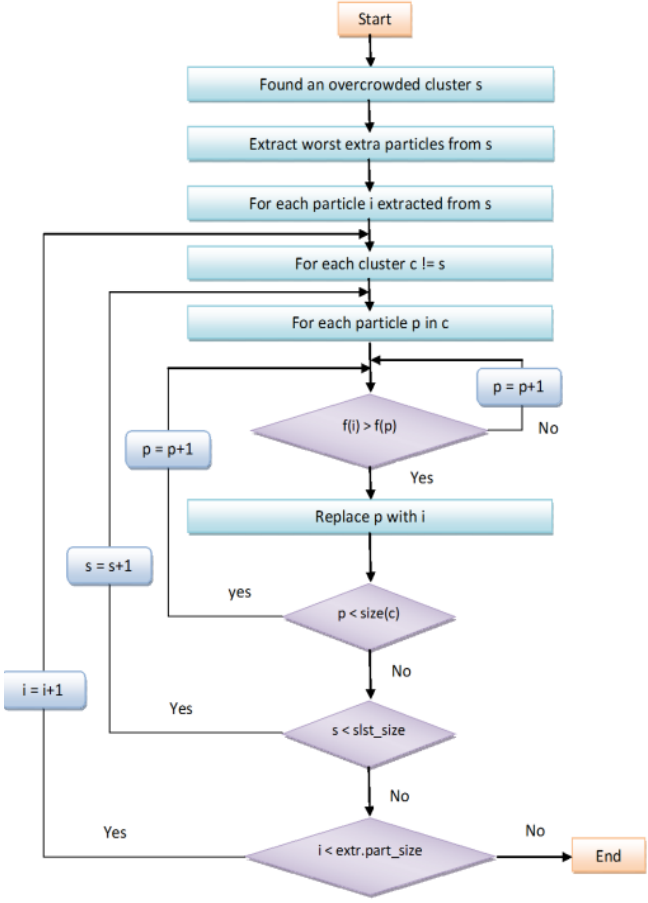

Fig 5. Applying overcrowding check for a cluster

$$F(x,t) = \max_{i=1...p} \frac{H_i(t)}{1+W_i(t)\sum_{j=1}^{D}(x_j(t)-X_{ij}(t))^2} \quad (4)$$

$H_i(t)$: height of peak i at time t
$W_i(t)$: width of peak i at time t
$X_{ij}(t)$: jth element of location of peak i
p: number of peaks.

Table 1: MPB Problem Settings

| Parameter Name | Parameter Value |
|---|---|
| **Number of dimensions, D** | 5 |
| **H** | [30,70] |
| **Shift length, s** | 1.0 |
| **Height severity** | 7.0 |
| **Width severity** | 1.0 |
| **Peak shape** | Cone |
| **p (no. of peaks)** | 10 |
| **$U_{cf}$ (change frequency)** | 10000 |
| **Basic function** | No |
| **Correlation coefficient, λ** | 0 |
| **S** | [0,100] |
| **W** | [1,12] |
| **I** | 50 |

The peak can move in any direction with velocity $v_i(t)$. This is given below:

$$v_i(t) = \frac{s}{|r+v_i(t-1)|}((1-\lambda)r + \lambda v_i(t-1)) \quad (5)$$

Where

  s: shift length, determines severity of problem
  r: random vector
  λ: correlated parameter.

A peak update its position, height and width in the following ways:

$$H_i(t) = H_i(t-1) * \text{height severity} * \sigma \quad (6)$$

$$W_i(t) = W_i(t-1) * \text{width severity} * \sigma \quad (7)$$

$$X_{ij}(t) = X_{ij}(t-1) + v_i(t) \quad (8)$$

Here σ normal distribution random number with mean 0 and variation of 1.
To calculate performance of algorithm, offline error is used.

$$\mu = \frac{1}{N}\sum_{n=1}^{N}(h_n - f_n)$$

Where

$h_n$ = best value of $n^{th}$ environment
N = total environments
$f_n$ = optimal solution of $n^{th}$ environment

We first find differences between best solutions and optimal solutions per environmental change then we find average of all these differences. Fitness evaluations are calculated by multiplying change frequency with total number of environments i.e.

  fitness evaluations = N * $U_{cf}$

For our experiments, we have set N = 100 and $U_{cf} = 10^6$ and average of 50 runs with different seeds is taken for these experiments.

## V. RESULTS

Different experiments have been performed with different configurations. MPB problems have been used for experiments with default settings. Different combinations of M and N are used where M denotes size of cradle swarm and N denotes size of sub-swarm. Table 2 shows offline error with average of 50 runs. Total numbers of generated clusters are shown in Table 3. Number of survived clusters are shown in Fig 4. Survived clusters contain both converged and non-converged clusters. Table 4 shows number of found peaks.

Table 2: Offline Error with different values of parameters

|         | N = 2 | N = 3 | N = 4 | N = 5 | N = 7 |
|---------|-------|-------|-------|-------|-------|
| M = 10  | 3.76  | 4.46  | 4.83  | 7     | 5.3   |
| M = 30  | 1.75  | 1.97  | 2.46  | 3.5   | 3.4   |
| M = 50  | 1.4   | **1.39** | 1.77 | 1.6  | 2.54  |
| M = 70  | 1.48  | **1.01** | 1.47 | 1.51 | 1.95  |
| M = 100 | 2.3   | 1.5   | **1.1** | 1.2 | 1.76  |

If a peak is within search radius of a cluster then we consider it to be found by algorithm. We have used this measure because it performs very well for our experiments i.e. when performance of algorithm is compared with different configuration values of M and N then it gives very good results.

Table 3: Number of Clusters generated

|         | N = 2 | N = 3 | N = 4 | N = 5 | N = 7 |
|---------|-------|-------|-------|-------|-------|
| M = 10  | 5     | 5     | 3     | 2.39  | 3     |
| M = 30  | 16    | 12.5  | 8     | 7     | 5.5   |
| M = 50  | 25.5  | 17    | 14    | 12    | 8.5   |
| M = 70  | 38    | 26.5  | 20    | 17    | 11    |
| M = 100 | 55    | 33    | 28    | 22    | 17    |

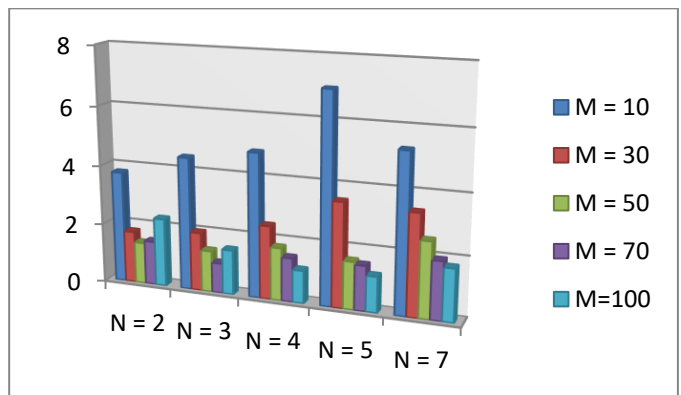

Fig. 6. Offline Error

It can be seen from Table 2 and Fig 6 that configuration plays very vital role in performance of DCPSO. If we set value of *M* very high or very low while value of *N* remains fixed, then performance of DCPSO also changes. It was observed from experiments that where size of cradle swarm *M* is set to 70 and maximum sub-swarm size is set to 30 then our algorithm gives optimum results.

Table 4: Number of Peaks Found

|         | N = 2 | N = 3 | N = 4 | N = 5 | N = 7 |
|---------|-------|-------|-------|-------|-------|
| M = 10  | 3.8   | 3.16  | 2.8   | 2.41  | 2.32  |
| M = 30  | 5.99  | 5.5   | 4.54  | 4.12  | 3.6   |
| M = 50  | 6.89  | 6.45  | 5.67  | 5.04  | 4.58  |
| M = 70  | 8     | 7.25  | 6.42  | 5.9   | 5.21  |
| M = 100 | 7.26  | 7.75  | 7     | 6.48  | 5.76  |

Analysis of Table 3 and Table 4 shows that DCPSO found more peaks when we set population size larger. In fact, when we set population size larger, then more clusters are generated, and more peaks are found by algorithm so performance of DCPSO also increases. But to achieve optimal results, we have to set change frequency according to problem otherwise we may get different results.

**Comparison of DCPSO and CPSO**

We have compared our results with latest clustering PSO for dynamic environment [20]. A comparison of CPSO and DCPSO is given in Fig 8. We have performed this experiment with default configurations setting the size of population 70 and the cluster size (max_size) equal to 3 for CPSO and DCPSO.

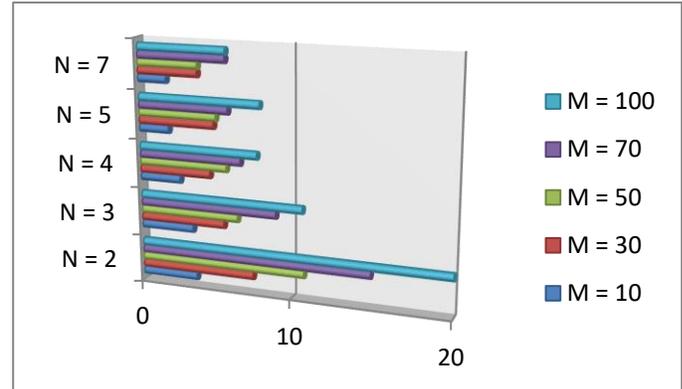

Fig. 7. Survived Clusters

We can see in Fig 8 that with these configuration setting our proposed algorithm performs better in all comparisons i.e. it generates more clusters, found more peaks, give less offline error and result in more number of survived sub swarms.

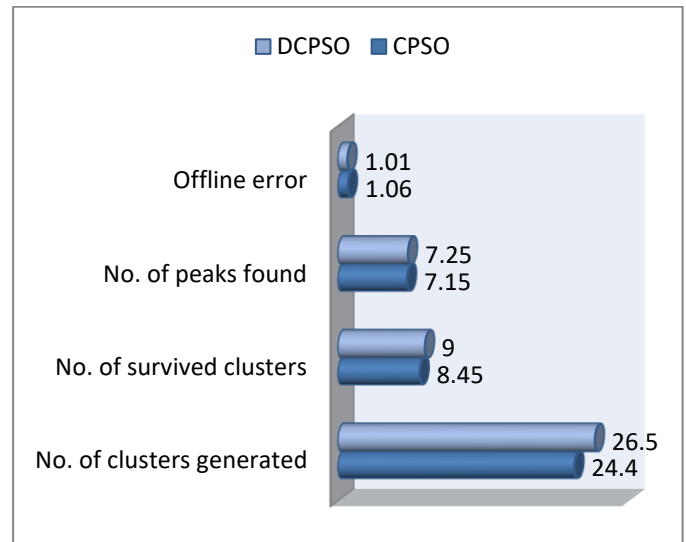

Fig 8. Comparison of CPSO and DCPSO

If we take careful analysis of this comparison and other experiments then we can see that these results also support our idea/concept that by using explore_area() method, our algorithm try to find out the undiscovered area and then exploit those area(s) to find more peaks in the search space. Although, there is no guarantee that it will always find an undiscovered area because it is also possibility that new created cluster may overlap with some existing cluster(s) so in this case we perform overlapping check. But, still, our algorithm try to perform better by increasing the diversity of the algorithm as we have discussed before that due to environment change and overlapping, overcrowding and convergence check, the population size and number of clusters may decrease and, in this situation, it becomes very difficult for any algorithm to locate and track multiple peaks. So, if our algorithm does not find a new location then, still, it tries to improve its performance.

By taking analysis of all experiments, it can be seen that performance of DCPSO vary if we change initial population size or if cluster size is changed. So, we have to set optimal configuration according to problem to achieve optimal results. It was seen that our algorithm gives optimal results when we set cradle swarm size to 70 and sub-swarm size to 3 with default configuration on MPB problem.

### A. Conclusion and Future work

Many Particle Swarm Optimizer algorithms are present currently for Dynamic Optimization problems. Many researchers have used multi swarm method to locate and track multiple optima in dynamic environment. But, there are also some issues which needs to be considered when using multi swarm method; For example, how clusters should be created, how particles of different clusters should be guided to divert them to a specific sub-region, how number of clusters should be determined and so on.

Our proposed DCPSO performs very well for dynamic environment by efficiently finding and tracking multiple peaks. For generating proper number of clusters, a single linkage hierarchical clustering scheme is used. Then to exploit the undiscovered sub regions and to move particles to specific sub regions, a local search method is used. Our algorithm also uses a different learning strategy to speed up searching and convergence process. Then our algorithm extracts the best information by exploring the undiscovered regions of search space to find a new best position and to improve the diversity of clusters by maintaining clusters strength. DCPSO also used a scheme to handle dynamic environment by using best positions of previous environment in new environment.

Based on experimental results, we can conclude that our proposed algorithm performs very well in terms of finding and tracking multiple optima in dynamic environment. We have compared performance of our algorithm with CPSO and found that our algorithm outperforms then CPSO w.r.t. change severity in search space.

Finally, according to experiments, DCPSO is a good optimizer for dynamic environments specially when there are multiple changing peaks in dynamic fitness landscape.

Although, DCPSO is performing very well still there are some issues which can be addressed in future. E.g. during convergence check, converged clusters are removed which results in reduction of population size. Although we add some particles randomly but still there is possibility that these particles may be attracted by existing clusters. So more work needs to be done.

We have set fixed value of sub-swarm size in our algorithm which is not a good approach so more work can be done to make it self-adaptive.